%% file: main.tex
\documentclass[letterpaper, conference]{ieeeconf}
\IEEEoverridecommandlockouts                                       
\usepackage[utf8]{inputenc}
\usepackage{amsmath,amssymb,amsfonts}
\usepackage{bm}
\usepackage[hyphens]{url}
\usepackage{graphicx,graphics}
\usepackage{multirow}
\usepackage{booktabs}
\usepackage{multirow}
\usepackage[ruled,linesnumbered]{algorithm2e}
\usepackage{textcomp}
\usepackage{tabularx}
\usepackage{relsize}
\usepackage[flushleft]{threeparttable}
\usepackage{cite}
\usepackage[table,dvipsnames]{xcolor}

\newcommand{\midsepremove}{\aboverulesep = 0.2mm \belowrulesep = 0.2mm}
    \midsepremove
\newcommand{\midsepdefault}{\aboverulesep = 0.605mm \belowrulesep = 0.984mm}
    \midsepdefault
\newcommand{\myfigureshrinker}{\vspace{-0.4cm}}
\DeclareMathOperator*{\argmin}{arg min}
\title{\LARGE \bf Grasp Transfer for Deformable Objects\\by Functional  Map  Correspondence}
\author{{Cristiana de Farias$^{{\dagger*}}$, Brahim Tamadazte$^{{\ddagger}}$, Rustam Stolkin$^{{\dagger}}$, Naresh Marturi$^{{\dagger}}$}
\thanks{This work was supported by the UK National Centre for Nuclear Robotics (NCNR). Part funded by CHIST-ERA under Project EP/S032428/1 PeGRoGAM and in part supported by Faraday Institution sponsored Recycling of Lithium Ion Batteries (ReLiB) project (grant: FIRG005).}%
\thanks{$^{\bm{\dagger}}$Extreme Robotics Laboratory, School of Metallurgy and Materials, University of Birmingham, Birmingham, United Kingdom. $^{\bm{\ddagger}}$Sorbonne Universit\'{e}, CNRS UMR 7222, INSERM U1150, ISIR, F-75005, Paris, France. $^{*}$Corresponding Author: {\tt CXM1029@student.bham.ac.uk} }%
}
\begin{document} \sloppy
\bstctlcite{IEEEexample:BSTcontrol}
\maketitle
\input{sections/abstract.tex}
%
\section{Introduction}
\label{sec:intro}

\input{sections/intro.tex}
\input{sections/overview}

\section{Grasp Transfer Through Functional Maps}
\input{sections/preprocessing}

\subsection{Functional Map Correspondence}
\label{sec:Functional_Maps}
\input{sections/functionalMap}

\subsection{Grasp Transfer}
\label{sec:Grasp_Transfer}
\input{sections/graspTransfer}

%
%

\section{Experimental Validations}
\label{sec:experiments}
\input{sections/experimentsSim}

%
\section{Conclusion}
\label{sec:conclusion}
\input{sections/conclusion.tex}
\typeout{}
\bibliographystyle{IEEEtran}
\bibliography{references}
\end{document}

%% file: sections/abstract.tex
\begin{abstract}
Handling object deformations for robotic grasping is still a major problem to solve. In this paper, we propose an efficient learning-free solution for this problem where generated grasp hypotheses of a region of an object are adapted to its deformed configurations. To this end, we investigate the applicability of functional map (FM) correspondence, where the shape matching problem is treated as searching for correspondences between geometric functions in a reduced basis. For a user selected region of an object, a ranked list of grasp candidates is generated with local contact moment (LoCoMo) based grasp planner. The proposed FM-based methodology maps these candidates to an instance of the object that has suffered arbitrary level of deformation. The best grasp, by analysing its kinematic feasibility while respecting the original finger configuration as much as possible, is then executed on the object. 
We have compared the performance of our method with two different state-of-the-art correspondence mapping techniques in terms of grasp stability and region grasping accuracy for 4 different objects with 5 different deformations. 
%
%
\end{abstract}%

%% file: sections/intro.tex
The past decade has seen unprecedented advances in robotic manipulation capabilities, propelled by advances in sensor and computing abilities. Nevertheless, reliable robotic grasping of non-rigid deformable objects remains an open and challenging research problem. The grasp planning literature predominantly focuses on searching for stable grasps over the surfaces of rigid objects\cite{adjigble2021spectgrasp,Adjigble2018LoCoMo, sahbani2012overview, SurveyMultifingerDeepLearning, de2021dual, deFarias2020BOGrasping}. Grasping deformable objects has so far received comparatively little attention\cite{sanchez2018robotic}. In this paper, we focus on the problem of adapting a known grasp of a deformable object, to find a corresponding stable grasp on a different configuration (\textit{i.e.}, a different deformation) of that object. In the sense that, subject to topology, any object shape can be warped to map onto any other object shape, this is related to the problem of transfer, of a known grasp on a known object, to a new grasp on a new object. In this light, we can consider a warped instance of a known object to be also equivalent to a transfer from a known object to a novel object \cite{Hillenbrand2012TransferringFunctional, Theodoros2015TransferringFunctional}.


%
%
The problem of transferring a grasp between two objects, can be formulated as transferring information between similar surface regions. Predominantly, works on this topic fall into two categories: extrapolating kinesthetic information to novel objects\cite{kopicki2016oneshotlearning,kopicki2019bettergrasps, Hjelm2014cross-tasktransfer}; and, exploring similarities between familiar objects to generalise grasps\cite{Vahrenkamp2016Part-Based,Tian2019TransferringReplanning, Theodoros2015TransferringFunctional, Hillenbrand2012TransferringFunctional, adjigble2019assisted}. In the former, approaches such as in\cite{kopicki2016oneshotlearning} and \cite{kopicki2019bettergrasps} use a demonstration to learn grasps and generalise them to novel objects while maintaining same grasping characteristics. Although these methods are efficient in synthesising grasps for unknown or novel objects, they majorly focus on local features without semantic meaning. Our work falls within the latter category, which, in contrast, trades-off the generalisation to focus on shape similarities and gain a richer understanding of the relations between objects.

\begin{figure}
    \centering
    \includegraphics[width=\columnwidth]{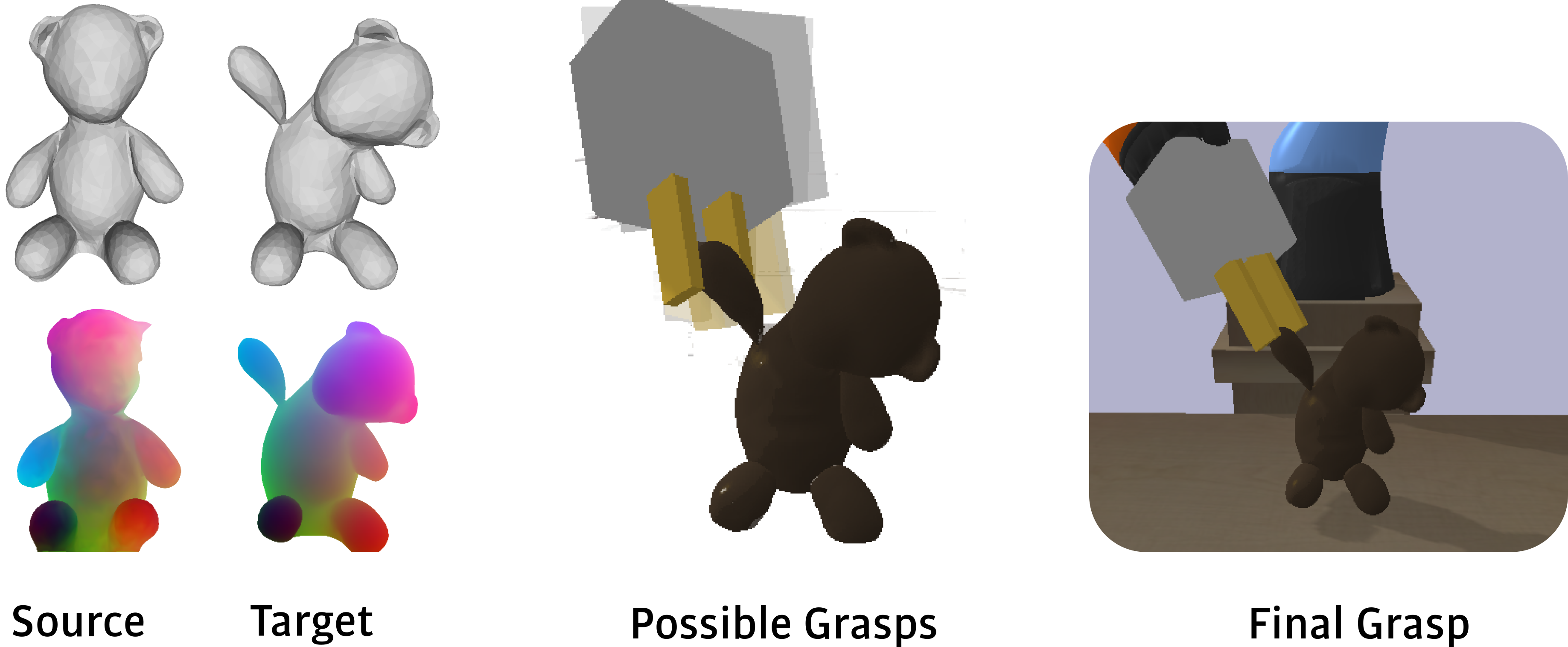}
    \caption{Deformable object grasping by the proposed grasp transfer method. (left) Source and target meshes, and their functional maps. (middle) Transferred possible grasp configurations. (right) Executed best grasp.}
    \label{fig:teaser}
\end{figure}

Some closely related works to ours explore grasp transfer between similar objects or of the same category~\cite{Stuckler2014nonrigid, rodriguez2018transferring, Rodriguez2018a, lin2018nonrigid, Theodoros2015TransferringFunctional, Hillenbrand2012TransferringFunctional, Tian2019TransferringReplanning}. Authors in ~\cite{Theodoros2015TransferringFunctional, Hillenbrand2012TransferringFunctional} have proposed a method to transfer grasps between same category objects by rigid alignment of similar shapes, contact wrapping, and local re-planning. Similarly, \cite{Tian2019TransferringReplanning} solved this problem by means of bijective contact mapping and grasp re-planning. In this case, rigid alignment was found by sampling the surface of two objects and minimising the deviation between points. A non-rigid registration method based on Coherent Point Drift (CPD) is used in \cite{Stuckler2014nonrigid} to transfer manipulation skills between objects. In \cite{lin2018nonrigid}, the authors have used CPD to transfer and refine grasps between the same class of objects. A learning method using CPD to account for shape deviations when transferring manipulation skills between objects is proposed in~\cite{Rodriguez2018a, rodriguez2018transferring}.  We note that, variations of CPD are the main choice for non-rigid shape matching in previous works. Albeit, an efficient choice, CPD is prone to get stuck in local minima and needs a good initial pose to work well; thus, leading to subpar results in the presence of larger deformations.

\begin{figure*}
    \centering
    \includegraphics[width=0.9\textwidth]{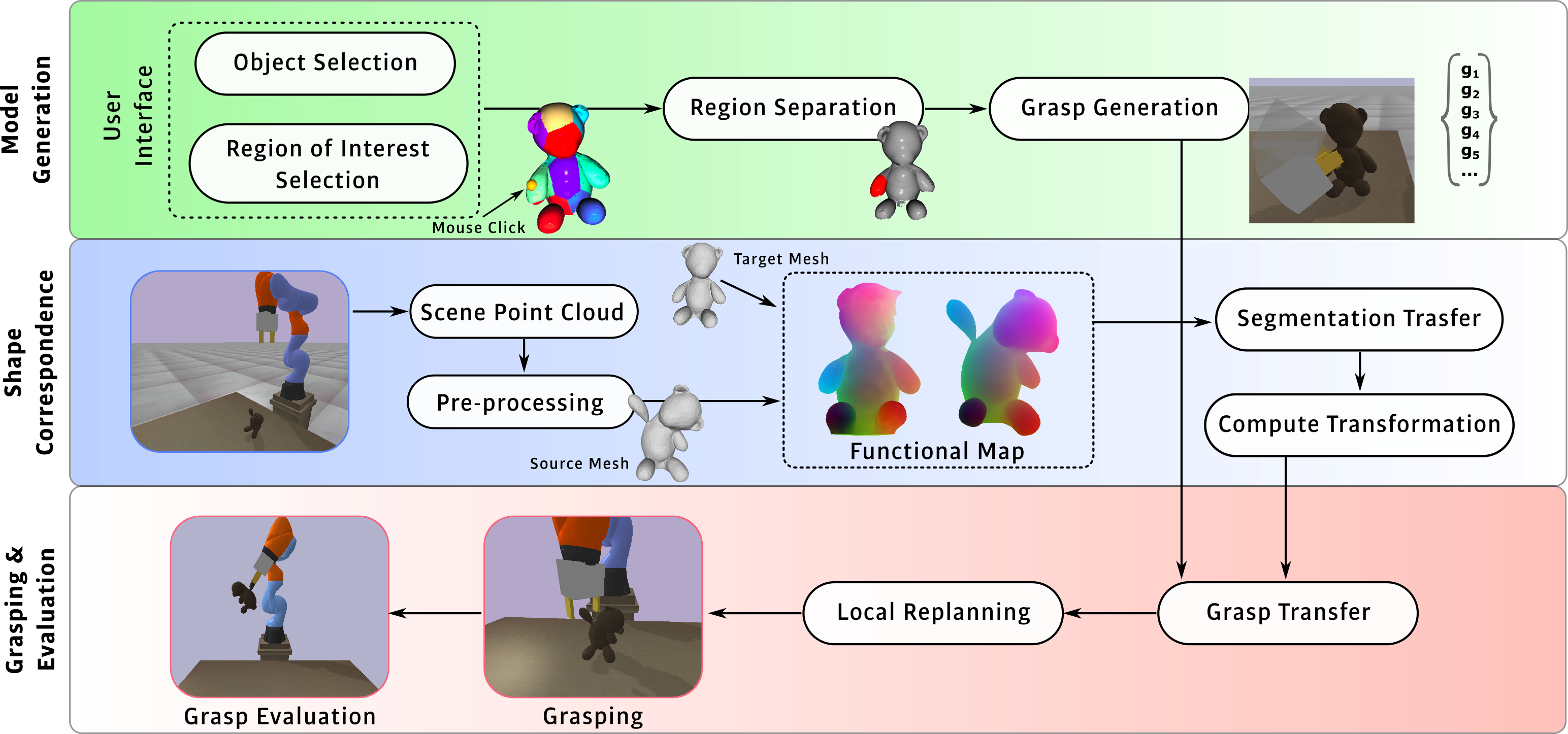}
    \caption{Proposed pipeline of our methodology to transfer non-rigid grasps using functional maps. It consists of three steps: grasp model generation (top -- green); finding shape correspondences (middle -- blue); and, grasp transfer (last -- red). }
    \label{fig:FMG_Flow}
    \myfigureshrinker
\end{figure*}

We believe that functional map (FM) correspondence, which was first formulated in \cite{OvsjanikovFMOriginal}, can be an efficient alternative to CPD for handling object deformations. 
Within the FM framework, the shape matching problem is treated as searching for correspondences between geometric functions in a reduced basis. This choice to treat point correspondences between objects as functions leads to simpler convex least-square optimisation problems and provide greater flexibility to incorporate linear constraints to the problem. FM with several applications in different areas of computer graphics such as partial matching, deformation and symmetry analysis, exploration of shape collections etc. \cite{Ovsjanikov2017Tutorial, sahilliouglu2020survey, Biasotti2016ShapeMatchingOverview}, has been shown to perform well in challenging settings \cite{SHREC2020Benchmark}. 

In this paper, we exploit the FM framework to propose a robust learning-free solution to the problem of transferring grasps between objects that have suffered non-rigid deformations. Despite having many interesting properties, FM has not been explored in robotics. To the best of our knowledge, this is the first work to apply the FM pipeline for robotic grasping. Principally, our solution leverages user input and grasps generated by our Local Contact Moments (LoCoMo) based grasp planner \cite{Adjigble2018LoCoMo} to create a ranked list of grasps focused on a region of interest of an object.  We then propose a FM-based method to map these grasps to an instance of the object that has suffered arbitrary level of deformation (see Fig. \ref{fig:teaser}). Finally, we transform and adapt the grasps to the deformed object and execute the best (kinematically feasible) grasp. We validate our method by performing a number of experiments, using a simulated 7-axis robot fitted with a 2-finger gripper, on different objects with multiple random deformations. We also compare and discuss our method's performance with two closest state-of-the-art techniques.

%% file: sections/overview.tex
\section{Problem Statement and Pipeline}
%
Our approach exploits FM to compute the correspondences between features of two shapes and then use this mapping to transfer the grasp configuration. We denote a shape by the Riemannian manifold embedded in $\mathbb{R}^3$. Let $\mathcal{X}, \mathcal{Y}$ be respectively the source and target shapes, with $n_{\mathcal{X}}$ and $n_{\mathcal{Y}}$ vertices stored in triangular meshes. ${T}_p:\mathcal{Y}\rightarrow\mathcal{X}$ represents the point-wise correspondence map. By solving the shape matching problem, we obtain the relation between each part of the object. This is used to find the optimal transformation $\mathbf{T}\in\text{SE}(3)$ that successfully transfers a grasp to the target object. %

The pipeline of our method is shown in Fig. \ref{fig:FMG_Flow}. It consists of three main steps: grasp model generation, shape correspondence, and grasp transfer and evaluation. 
Initialised by the user input, we generate grasping models to be applied on segmented object region. Later, we find the relation between source $(\mathcal{X})$ and target $(\mathcal{Y})$ object shapes. FM framework helps us in accomplishing this shape correspondence. 
Finally, we transfer and adjust the grasp to the target so that it is both stable and feasible. We evaluate the grasp stability as in \cite{Bekiroglu2020Benchmark}.
%
%
%
 %
%
%
 %

%% file: sections/preprocessing.tex
Below, we present our methodology of grasp transfer to handle deformable objects. We start with our data acquisition process and present in detail the steps mentioned in pipeline.

\subsection{Data Acquisition}\label{sec:dataacq}
\label{sec:preprocessing}In this work, we focus only on full and close isometric shape matches for grasp transferring. Hence, it is necessary to ensure both source and target shapes remain consistent.%

Let $\mathcal{C}_\text{obj}$ be the ground plane segmented point cloud of an object, which is obtained by stitching multiple point clouds acquired from different positions. It is worth noting that these clouds are acquired by a camera fixed on a robot whose transformation with the base frame is known; hence, all the clouds are in a common coordinate frame (robot base) suitable for stitching \cite{marturi2019dynamic}. Source and target clouds are represented by $\mathcal{C}_\mathcal{X}$ and $\mathcal{C}_\mathcal{Y}$, respectively. %
Next, we convert these point clouds to meshes using Poisson surface reconstruction algorithm ~\cite{kazhdan2013screenedPoisson}. This yields watertight meshes from oriented point sets while preserving geometric details. 
As obtained meshes are generally very dense we add an extra step of quadric mesh decimation to simplify it while preserving the original mesh shape, volume, and boundaries.

\subsection{Grasp Model Generation}\label{subsec:GraspModel}

The grasp model is a set of variables needed to fully describe grasps with respect to a shape. If the problem is simplified, \textit{i.e.}, to transfer grasps between same rigid object at different poses, then we only need a list of grasp hypothesis and an object model to define the grasping model. The transferring problem now becomes finding an optimal rigid transformation between the source (or model) point cloud and the target. 

\begin{figure}
    \centering
    \includegraphics[width=1\columnwidth]{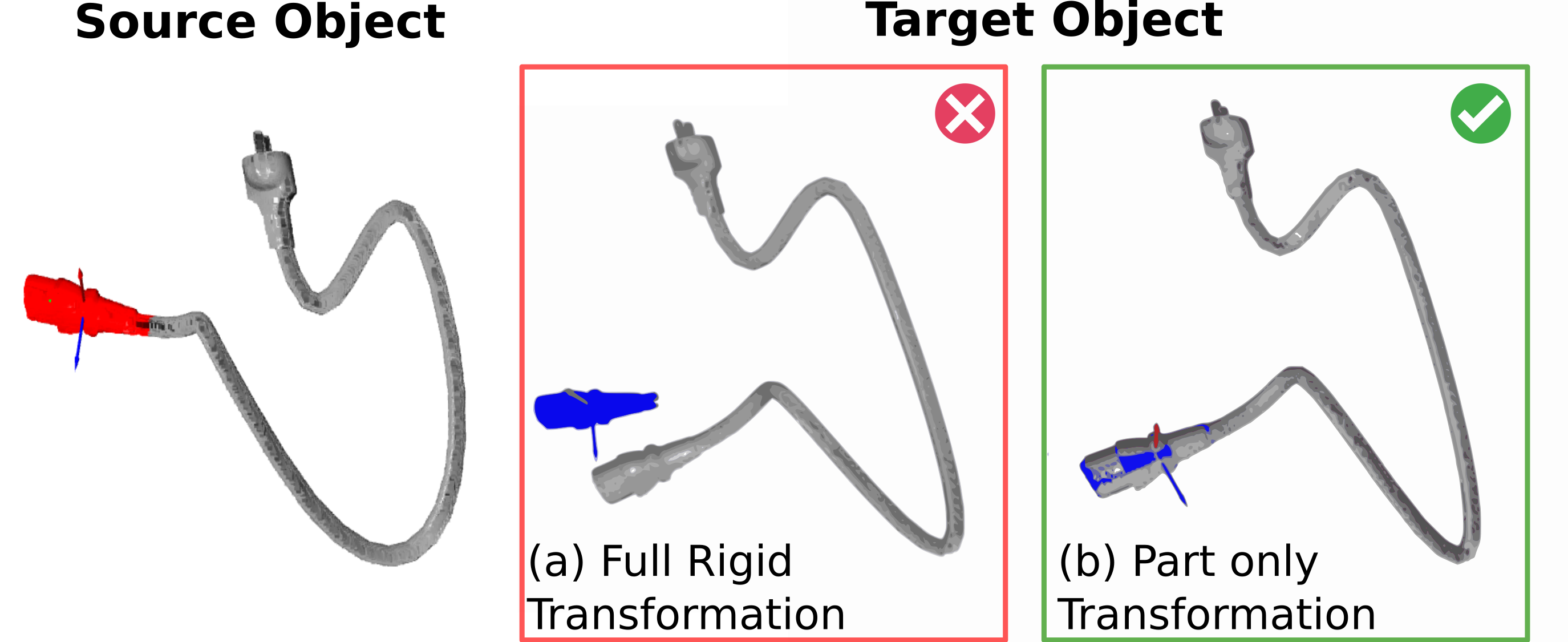}
    \caption{Transformation of a region (in red) of the source object (power cable). (a) By applying the optimal transformation between all points (full object rigid pose) and (b) by finding the transformation of the points inside the segmentation only (applying local region transformations).}
    \label{fig:Grasp Segmentation}
\end{figure}

However, transferring grasps between non-rigid shapes is a more complex problem. Even if the shapes are isometric (\textit{i.e.}, preserve their geodesic distances), the rigid transformation suffered by different parts of the object might not be constant, and most likely will be non-linear. In order to alleviate this, we segment the object into multiple parts and consider the transferring problem as obtaining rigid transformations for these segmented regions. This is shown in Fig. \ref{fig:Grasp Segmentation}. We assume that the segmented object parts are big enough to contain all finger contacts of at least one grasp and that there are no significant deformations within the segmentation region. 
%

\begin{figure}
    \centering
    \includegraphics[width=\columnwidth]{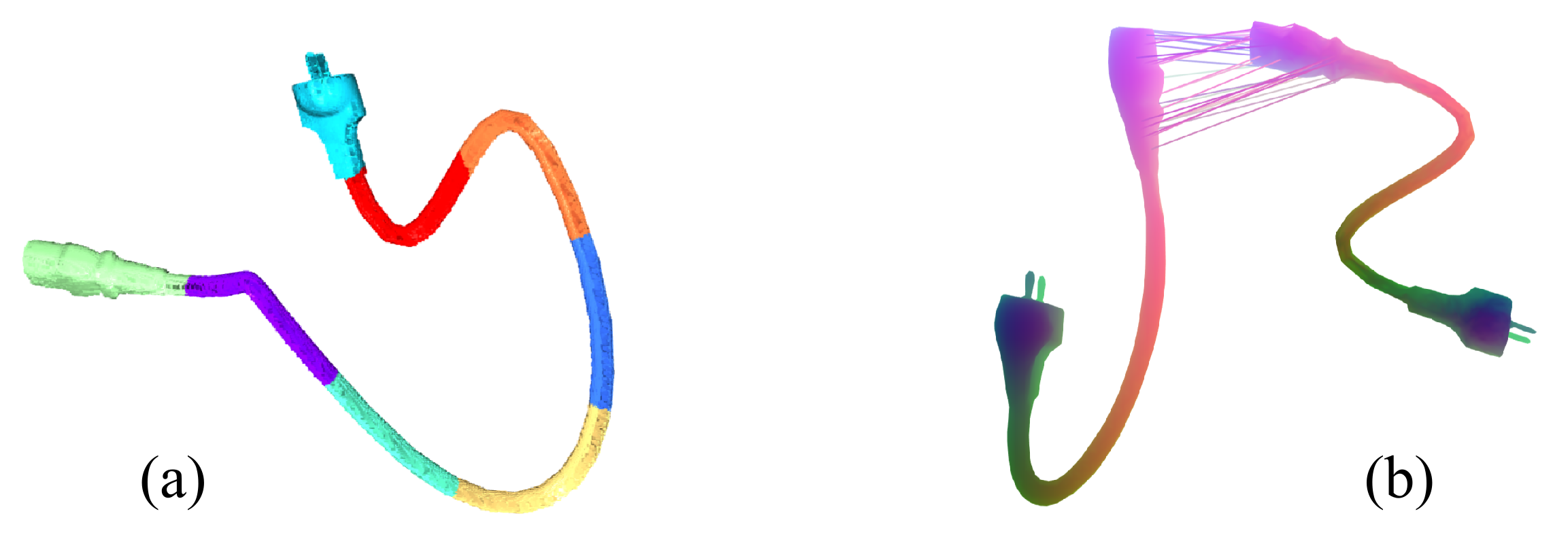}
    \caption{(a) Object segmented into different parts using \textit{k-means} clustering, each colour denotes a segment. User will select the segment where the robot will grasp the object. (b) Point-to-point correspondence between two objects. Similar colours indicate the same region in both objects.} 
    \label{fig:ClustersSegmentation}
\end{figure}

Generally, when grasping an object, our aim is to perform some task with it. Thus, grasp should be planned in a way that it supports task accomplishment. Previously, some works dealing with this task-aware grasping have followed learning-based methodologies where affordances are assigned to different object parts depending on the desired task~\cite{Song2015TaskBasedPlanning,Madry2012ProbabilisticTransfer,Hjelm2014cross-tasktransfer}. 
In this work, our approach is by leveraging human knowledge of how objects should be grasped. 
For this, a user interface is created to select an initial region of interest on the object. 
The initial interface will present the user with a segmented cloud of an object as shown in Fig. \ref{fig:ClustersSegmentation}(a) Segmentation is performed by partitioning the point cloud into $N_C$ parts using \textit{k-means} clustering. 
The segmented cloud is denoted as $\mathcal{S}_\mathcal{X}$ with $\mathcal{S}_\mathcal{X}\subseteq \mathcal{C}_\mathcal{X}$;

In this work, we synthesise the grasps on $\mathcal{C}_\mathcal{X}$ using our LoCoMo grasp planner \cite{Adjigble2018LoCoMo}. It generate grasps based on local similarity between an object surface and gripper fingers using zero moment shift. Grasp poses are provided in a ranked list based on shape similarity score. We denote the ranked list of $N_{G}$ grasps poses in the world frame as the set $\mathcal{G}^0=\{\mathbf{g}^0_i \in \text{SE}(3), i=1...N_{G}\}$. As we are considering only part based grasps, we filter out the grasps leading to finger placements outside the segmented region, such that $\mathcal{G}^0_{S_\mathcal{X}}\subseteq\mathcal{G}^0$ is our subset of grasp poses in the world frame. 

Finally, after segmentation and grasp generation we define a grasp model as the set $M=\left\{\mathcal{G}^0_{S_\mathcal{X}}, \mathcal{S}_\mathcal{X}, \mathcal{X} \right\} $, with $\mathcal{X}$ being the full surface of the source object, reconstructed from $\mathcal{C}_\mathcal{X}$.

%% file: sections/functionalMap.tex
Given two shapes (Riemannian manifolds), ${\mathcal{X}}, {\mathcal{Y}}$ with $n_{\mathcal{X}}$ and $n_{\mathcal{Y}}$ vertices, the surface correspondence problem generally aims at finding the bijective transformation $T_p:{\mathcal{Y}}\rightarrow {\mathcal{X}}$, which maps the vertices of ${\mathcal{Y}}$ onto $ {\mathcal{X}}$. This problem, given that  $n_{\mathcal{X}} = n_{\mathcal{Y}}$, can also be formulated as finding the permutation matrix $\mathbf{\Pi}$ of size $n_{\mathcal{Y}}\times n_{\mathcal{X}}$ in which all lines will sum up to 1. Although intuitive, inferring $\mathbf{\Pi}$ directly can be a complex problem which lacks flexibility and is badly-suited for more general rigid deformations. FM is an alternative technique that has been widely adopted for solving this over the last few years. 
The underlying idea is that it is often easier to optimise between real-valued functions rather than points in a shape. Indeed, finding matches in 3D space can lead to complex non-convex optimisation problems, whereas FM offers an elegant formalism that allows for a compact matrix representation in a low rank basis. Furthermore, this method allows for the easy incorporation of linear constrains to regularise the map, which yields simple convex least squares optimisation problems that are much more tractable~\cite{OvsjanikovFMOriginal}. For a more detailed understanding, we refer the reader to~\cite{Ovsjanikov2017Tutorial}. 
 
We define $f:\mathcal{X}\rightarrow\mathbb{R}$ and $h:\mathcal{Y}\rightarrow\mathbb{R}$ as real functions over the shapes. Then, we can use $T_p$ to transfer $f$ to $\mathcal{Y}$ via the composition $h=f\circ T_p$, which yields that $h(p)=f\left ( T_p(p) \right )$ for any $p\in\mathcal{Y}$. As the mapping between functions is linear, we can represent the transformation in matrix form, $\mathbf{h}=\mathbf{\Pi}\mathbf{f}$, with $\mathbf{f}$ and $\mathbf{h}$ being the vector forms of their counterpart functions and $\mathbf{\Pi}$ being any well defined linear map between functions (not necessarily a bijection).

Another important aspect of FM is to use reduced basis for the functions instead of working in the full $\mathbb{R}^{n_{\mathcal{Y}}}$ or $\mathbb{R}^{n_{\mathcal{X}}}$ spaces. Thus, we compute a reduced set of $k_\mathcal{X}, k_\mathcal{Y}$ basis functions over $\mathcal{X}$ and $\mathcal{Y}$, and encode their coefficients as the columns of $\mathbf{\Phi}_\mathcal{X}$ and $\mathbf{\Phi}_\mathcal{Y}$, which are the basis over $\mathcal{X}$ and $\mathcal{Y}$, respectively. We note that the number of basis in which we describe our correspondence in this formulation is usually less than the number of vertices in the object, \textit{i.e.}, $k_\mathcal{X}\ll{n_{\mathcal{X}}}$ and $k_\mathcal{Y}\ll{n_{\mathcal{Y}}}$. Furthermore, for $\mathbf{f}_1,\mathbf{f}_2\in \mathcal{X}$ we define the inner product on the manifold as $\langle \mathbf{f}_1,\mathbf{f}_2\rangle_{\mathcal{X}} = \mathbf{f}^T_1\mathbf{A}_\mathcal{X}\mathbf{f}_2 $, with $\mathbf{A}_\mathcal{X}$ being a diagonal matrix of weights (and equivalently, for $\mathbf{h}_1,\mathbf{h}_2\in \mathcal{Y}$, $\langle \mathbf{h}_1,\mathbf{h}_2\rangle_{\mathcal{Y}} = \mathbf{h}^T_1\mathbf{A}_\mathcal{Y}\mathbf{h}_2$). We highlight here that if the vectors are orthonormal with respect to the weighted inner product, then the inner product is an identity matrix. Thus, for orthonormal basis,$\mathbf{\Phi}_\mathcal{X}^T\mathbf{A}_\mathcal{X}\mathbf{\Phi}_\mathcal{X}=\mathbf{I}$ (or  $\mathbf{\Phi}_\mathcal{Y}^T\mathbf{A}_\mathcal{Y}\mathbf{\Phi}_\mathcal{Y}=\textbf{I}$). Equipped with this, 
we can write our functional transformation as
\begin{equation}
\begin{aligned}
    \mathbf{\Phi}_\mathcal{Y}\mathbf{C}&=\mathbf{\Pi}\mathbf{\Phi}_\mathcal{X}\\
    \mathbf{C}&=\mathbf{\Phi}_\mathcal{Y}^T\mathbf{A}_\mathcal{Y}\mathbf{\Pi}\mathbf{\Phi}_\mathcal{X},
\end{aligned}
    \label{eq:FM_System}
\end{equation}
with $\mathbf{C}$ being the FM matrix representation, which fully encodes the original map $T_p$. In \eqref{eq:FM_System} we assume that the initial correspondence is known. However, the shape matching problem consists of actually computing the value of $\mathbf{\Pi}$. 

Overall the FM pipeline followed in this work is summarised as follows:
 \begin{enumerate}
     \item[i)] Compute a set of orthonormal basis $\mathbf{\Phi}_\mathcal{X}$ and $\mathbf{\Phi}_\mathcal{Y}$. Here, we compute our basis from  the first $n$ eigen-functions provided by the Laplace-Beltrami Operator (LBO) \cite{OvsjanikovFMOriginal}. These eigen-functions are often referred to as the harmonic of the manifold. We note that this choice of basis is particularly well-suited for shape matching problems as it is invariant to isometries, rigid motions, and is easy to compute. Also, it provides a natural multi-scale way to approximate functions. In this work, we compute the LBO as explained in \cite{meyer2003discreteLBO}.
     \item[ii)] Compute a set of descriptor vector functions $\mathbf{f}_i\in \mathbf{\Phi}_\mathcal{X}$ and $\mathbf{h}_i\in \mathbf{\Phi}_\mathcal{Y}$ such that the unknown FM satisfies $\mathbf{C}\mathbf{f}_i\approx \mathbf{h}_i$. For this, we use the wave kernel signature (WKS) descriptor~\cite{Mathieu2011WKS}, which was proven efficient for
     a variety of datasets~\cite{Jing2028OrientationPreservingMaps}. It is invariant to isometry and robust to some non-isometric deformations.
     \item[iii)] Estimate an optimal value of $\mathbf{C}$ by solving a constrained least squares optimisation problem. $\mathbf{C}$ is usually optimised by minimising the descriptor preservation energy $ E(\mathbf{C})=\|\mathbf{C}\mathbf{F}-\mathbf{H}\|^2$, with ($\mathbf{F}$ and $\mathbf{H}$ being the matrices encoding the descriptors). However, a number of other linear constraints can be incorporated to the optimisation. As in \cite{Jing2028OrientationPreservingMaps}, we induce our maps to be approximately isometric, to be associated with point-to-point maps and for preserving orientations.
     \item[iv)] Convert the FM to the point-to-point correspondence vector $T_{\mathcal{X}\mathcal{Y}}$. In order to find the transformation between different parts of the object we need to recover the point-wise map from our functional map. This can be done by iterative alignment and refinement in the basis domain. To this extent, we use the Bijective ICP (BCIPC) algorithm \cite{Jing2028OrientationPreservingMaps}. In Fig. \ref{fig:ClustersSegmentation}(b) we show the point-to-point correspondences between two objects, where similar colours indicate similar regions.
 \end{enumerate}

%% file: sections/graspTransfer.tex
After recovering the point-to-point map $T_{\mathcal{X}\mathcal{Y}}$ between the two shapes we need to calculate the grasp transformation and ensure it is stable. To this end, we first calculate the optimal transformation $\mathbf{T}$ between the segmented regions in the two shapes. %
For this, we start by computing the $n$ vertices in $\mathcal{X}$, which correspond to the region defined by $\mathcal{S}_\mathcal{X}$ and storing their position in the $3 \times n$ matrix $\mathbf{V}_\mathcal{X}$. Given the map $T_{\mathcal{X}\mathcal{Y}}$, we can easily find the points corresponding to the segmentation region in $\mathcal{Y}$ and store their position in $\mathbf{V}_\mathcal{Y}$ such that the row $i$ in $\mathbf{V}_\mathcal{X}$ will correspond to the row $i$ in $\mathbf{V}_\mathcal{Y}$ for all $1\leq i\leq n$. Once $\mathbf{V}_\mathcal{X}$ and $\mathbf{V}_\mathcal{Y}$ are defined, we can calculate the optimal rigid rotation and translation, $(\mathbf{R}\in SO(3),\mathbf{t}\in \mathbb{R}^3)$ between the two sets of points by noting that $\mathbf{R}\mathbf{V}_\mathcal{X}+\mathbf{t}=\mathbf{V}_\mathcal{Y}$, and therefore, solving
\begin{equation}
    \mathbf{R, t} = \argmin_\mathbf{R, t} \sum^n \|\mathbf{R}\mathbf{V}^{(i)}_\mathcal{X}+t-\mathbf{V}^{(i)}_\mathcal{Y} \|^2
    \label{eq:transformation}
\end{equation}
Using  $(\mathbf{R},\mathbf{t})$ we get the rigid transformation $\mathbf{T}\in SE(3)$. %
Now, the transferred grasp can be calculated as $\mathbf{g}_\mathcal{Y} = \mathbf{T}\mathbf{g}_i$, where $\mathbf{g}$ is the top-ranked feasible grasp in $\mathcal{G}_{\mathcal{S}_\mathcal{X}}$. In order to check if the new grasp is feasible we perform a inverse kinematics (IK) check after the transformation is applied. 

For the simplest scenario, in which our map inside the segmentation is perfect and both shapes correspond perfectly \textit{i.e.}, there are no deformations, the transformation $\mathbf{T}$ is enough to transfer the grasp between objects. However, this is not usually the case, and we need to take further steps to ensure our grasping is successful. To this aim, we introduce an extra step of local re-planning inspired from \cite{Tian2019TransferringReplanning}.

\begin{algorithm}[t]
	\SetAlgoLined
    \SetKwInOut{Input}{input}\SetKwInOut{Output}{output}
	\Input{$M$, $T_{\mathcal{X}\mathcal{Y}}$}
    \Output{$\mathbf{g}'_\mathcal{Y}$}
    Get $\mathbf{V}_\mathcal{X}$ from $\mathcal{S}_\mathcal{X}$, $\mathcal{X}$ \\
    Get $\mathbf{V}_\mathcal{Y}$ from $T_{\mathcal{X}\mathcal{Y}}$, $\mathcal{Y}$ \\
    Calculate $\mathbf{R,t}$ from \eqref{eq:transformation} and get $\mathbf{T}$ \\
    \While{not IK }
    {
    Get next $\mathbf{g}_i$ from $\mathcal{G}_{\mathcal{S}_\mathcal{X}}$ \\
    Calculate $\mathbf{g}_\mathcal{Y} = \mathbf{T}\mathbf{g}_i$ \\
    Check if $\mathbf{g}_\mathcal{Y}$ is IK \\
    }
    Get $\mathbf{p}^{(j)}_{\mathcal{X}}$ for all $j$ from $\mathbf{g}_i$\\
    Get $\mathbf{p}^{(j)}_{\mathcal{Y}}$ from $T_{\mathcal{X}\mathcal{Y}}$, $\mathbf{p}^{(j)}_{\mathcal{X}}$\\
    Calculate $\mathbf{g}'_\mathcal{Y}$ from \eqref{eq:local_replanning}\\
    Grasp Object\\
	\caption{Grasp transfer for deformable objects.}
	\label{alg:Grasp Transfer}
\end{algorithm}
For a gripper with $N_F$ fingers, let $\mathbf{p}^{(j)}_{\mathcal{X}}$, $j=1...N_F$ be the point in the surface closest to the j-th finger when the griper is at $\mathbf{g}_i$. Then, with  $T_{\mathcal{X}\mathcal{Y}}$ we compute the equivalent points $\mathbf{p}^{(j)}_{\mathcal{Y}}$ in $\mathcal{Y}$. Furthermore, let $\mathbf{p}^{(j)}_{c}$ be fingers when the grasp is at $\mathbf{g}'_\mathcal{Y}$. Then, as we aim at $\mathbf{p}^{(j)}_{c}\approx \mathbf{p}^{(j)}_{\mathcal{Y}}$, the grasp can be re-planed by minimising
\begin{equation}
    \mathbf{g}'_\mathcal{Y} = \argmin_{\mathbf{g}'_\mathcal{Y}} \sum^{N_F}_{j=1} (\mu_1\|\mathbf{p}^{(j)}_{c}- \mathbf{p}^{(j)}_{\mathcal{Y}}\|)+\mu_2\epsilon+ \Psi,
    \label{eq:local_replanning}
\end{equation}
with $\mu_1$ and $\mu_2$ being scalar weights, $\epsilon$ being the error between  $\mathbf{g}'_\mathcal{Y}$ and  $\mathbf{g}_\mathcal{Y}$ (indicating how much the grasp has changed) and $\Psi$ being a scalar penalty term when grasps are unreachable.

Our method is summarised in Algorithm \ref{alg:Grasp Transfer}.

%% file: sections/experimentsSim.tex
In this section, we detail the simulation experiments performed to validate the suitability of our approach that uses functional map correspondence for grasp transfer between objects that have suffered arbitrary level of deformation. We first present the details regarding the generation of deformed object models for validations and then discuss the method evaluations comparing its performance with two state-of-the-art methods, CPD\cite{myronenko2010point} and Iterative Closest Point (ICP)\cite{besl1992ICP}.%
\subsection{Object Deformation Dataset for Validations}
\begin{figure}
    \centering
    \includegraphics[width=0.9\columnwidth]{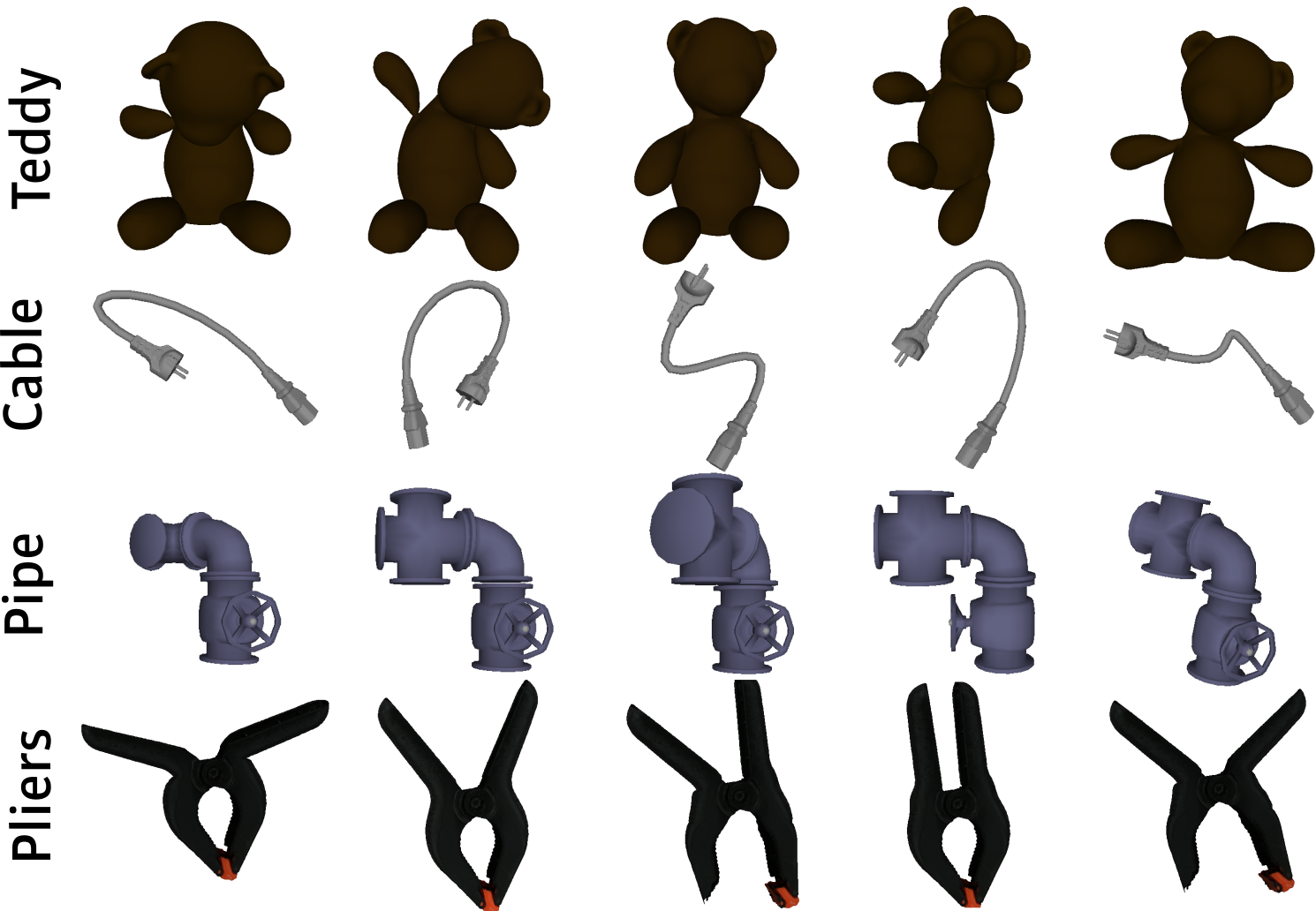}
    \caption{Objects and their sample deformations used in this work.}
    \label{fig:objects}
\end{figure}
Due to the lack of availability of datasets with multiple deformable models to be used for our experiments, in this work, we have produced a dataset of different objects featuring arbitrary non-rigid deformations. Currently, our dataset consists of 4 different object models and for each object it contains 5 different deformations. Treating each deformed pose as a new object, in total, it can be counted as 20 different objects. The initial 3D model of the object is either generated from multiple scans or downloaded from a 3D model database (e.g., power cable). These are then pre-processed (smoothing) and by using a 3D modelling software (Blender) we generated their deformed versions. The deformed models are exported as Wavefront object files, which can be loaded directly for experiments. Fig. \ref{fig:objects} shows generated object models.  Our dataset can be downloaded here: \url{https://gitlab.com/cristianafar/deformable-object-dataset}.
%

\subsection{Experimental Setup}
All simulation experiments are performed using the open-source PyBullet python module\cite{coumans2016pybullet}. 
Our robotic setup consists of a 7-axes robot arm fitted with a parallel jaw gripper. This robot and gripper configurations match with the kinematic models of KUKA iiwa robot arm and Schunk PG 70 gripper. For the sake of capturing scene data, we have simulated a 3D camera fixed on the end-effector of the robot. As our method needs good quality meshes, for each new object (scene), we have recorded the point cloud data by moving the robot to 15 different positions. As mentioned earlier in Sec. \ref{sec:dataacq}, we stitch these clouds together to generate input source mesh. To maintain a reasonable trade-off with the runtime, we have considered 3000 vertices for our source mesh. Next, we have implemented our FM correspondence using the package provided with \cite{Jing2028OrientationPreservingMaps}. For the FM we have used $50$ WKS descriptors and $100$ eigen-vectors as the LBO basis. In the grasping pipeline, $\mu_1=0.2$ and $\mu_2=0.8$ are set empirically. Grasping results are reported following the evaluation protocol presented in \cite{Bekiroglu2020Benchmark}.
\subsection{Grasp Transfer Evaluation}
\begin{figure}
    \centering
    \includegraphics[width=0.9\columnwidth]{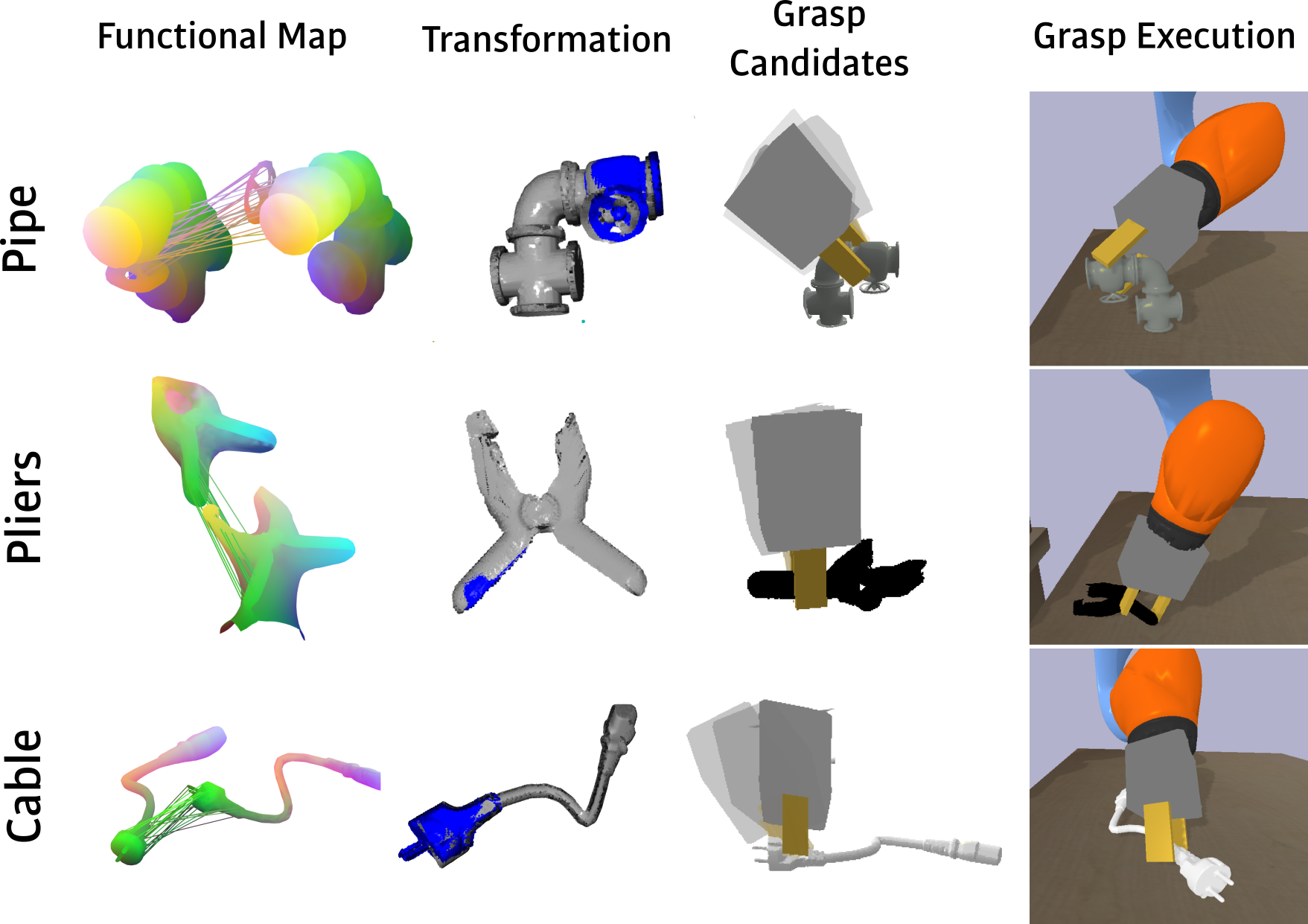}
    \caption{Illustration of grasp transfer with our proposed method. Detailed results can be found in the supplementary video.}
    \label{fig:FMPointMap}
\end{figure}
\midsepremove
\begin{table*}
\smaller
    \caption{Performance evaluation of the proposed grasp transfer method.}
    \label{tab:GraspEvaluation}
    \centering
    \begin{threeparttable}
    \begin{tabularx}{\textwidth}{>{\hsize=0.25\hsize\centering\arraybackslash}X|
                             >{\hsize=0.1\hsize\centering\arraybackslash}X
                             >{\hsize=0.1\hsize\centering\arraybackslash}X
                              >{\hsize=0.1\hsize\centering\arraybackslash}X
                              >{\hsize=0.1\hsize\centering\arraybackslash}X|
                             >{\hsize=0.1\hsize\centering\arraybackslash}X
                             >{\hsize=0.1\hsize\centering\arraybackslash}X
                             >{\hsize=0.1\hsize\centering\arraybackslash}X
                             >{\hsize=0.1\hsize\centering\arraybackslash}X|
                              >{\hsize=0.1\hsize\centering\arraybackslash}X
                             >{\hsize=0.1\hsize\centering\arraybackslash}X
                             >{\hsize=0.1\hsize\centering\arraybackslash}X
                             >{\hsize=0.1\hsize\centering\arraybackslash}X}
\toprule
    \multicolumn{1}{c|}{\textbf{Object}} & \multicolumn{4}{c|}{\textbf{Ours (FM-based method)}} & \multicolumn{4}{c|}{\textbf{CPD}} & \multicolumn{4}{c}{\textbf{ICP}} \\
    \cmidrule{2-13}
      {\textbf{Configuration\tnote{3}}} &  \textbf{Lift\tnote{1}} & \textbf{Rot.\tnote{1}} & \textbf{Shake\tnote{1} } &\textbf{Part\tnote{2}}  &  \textbf{Lift\tnote{1}} & \textbf{Rot.\tnote{1}} & \textbf{Shake\tnote{1} } &\textbf{Part\tnote{2}}  & \textbf{Lift\tnote{1}} & \textbf{Rot.\tnote{1}} & \textbf{Shake\tnote{1} } &\textbf{Part\tnote{2}}  \\
     \midrule
     \multicolumn{13}{c}{\textbf{Power Cable (Selected region: Plug)}}\\
     \midrule
     C1 & 100 & 100 & 100 & 100 & 100 & 100 & 100 & 100 & 0   & 0 & 0 & 0\\
     C2 & 100 & 100 & 100 & 100 & 0   & 0   & 0   & 0   & 0   & 0 & 0 & 0\\
     C3 & 100 & 100 & 100 & 100 & 100 & 100 & 100 & 100  & 100 & 0 & 0 & 0\\
     C4 & 100 & 100 & 100 & 100 & 100 & 0   & 0 & 0   & 0   & 0 & 0 & 0\\
     \midrule
     \multicolumn{13}{c}{\textbf{Teddy Bear (Selected region: Right Arm) }}\\
     \midrule
     C1 & 100 & 100 & 100 & 100 & 100 & 0  & 0 & 0 & 0   & 0   & 0   & 0\\
     C2 & 100 & 100 & 100 & 100 & 0   & 0  & 0 & 0 & 0   & 0   & 0   & 0\\
     C3 & 100 & 100 & 100 & 100 & 0   & 0  & 0 & 0 & 0   & 0   & 0   & 0\\
     C4 & 100 & 100 & 100 & 100 & 100 & 0  & 0 & 0 & 0   & 0   & 0   & 0\\
     \midrule
     \multicolumn{13}{c}{\textbf{Modular Pipe (Selected region: Wheel Section)}}\\
     \midrule
     C1 & 100 & 100 & 0   & 100 & 100 & 100  & 100 &  100 & 100 & 100 & 100 & 100\\
     C2 & 100 & 100 & 100 & 100 & 0   &  0   & 0   &  100 & 100 & 0   & 0   & 100\\
     C3 & 100 & 100 & 0   & 100 & 0   &  0   & 0   &  100 & 0   & 0   & 0   & 0  \\
     C4 & 100 & 100 & 100 & 100 & 100 &  0   & 0   &  100 & 0   & 0   & 0   & 100\\
     \midrule
     \multicolumn{13}{c}{\textbf{Pliers (Selected region: Left Handle)}}\\
     \midrule
     C1 & 100 & 100 & 100 & 0   & 100 & 100  & 100 & 0 & 100 & 100 & 0   & 0  \\
     C2 & 100 & 100 & 100 & 100 & 100 & 100  & 100 & 100 & 0   & 0   & 0 &100\\
     C3 & 100 & 0   & 0   & 100 & 100 & 100  & 100 & 100 & 100 & 100 & 100 &100\\
     C4 & 100 & 100 & 100 & 100 & 100 & 100  & 100 & 100 & 0   & 0   & 0   & 0  \\
     \midrule
  \textbf{Total} & \textbf{100} & \textbf{93.75} & \textbf{81.25} & \textbf{93.75} & 68.75 & 43.75  & 43.75 & 56.25 & 31.25   & 18.75  &  12.5  & 31.25  \\

     \bottomrule
\end{tabularx}
    \begin{tablenotes}
        \item[1] Percentage of success $\left(\bm{\%}\right)$ of lifting, Rotational and Shaking tests -- reported out of 3 trials for each configuration.
        \item[2] Percentage of success in grasping the correct object part -- Accuracy test.
        \item[3] Out of five models in our dataset, four deformation configurations of each object are used for robotic experiments.
    \end{tablenotes}
    \end{threeparttable}
    \myfigureshrinker
\end{table*}

For the sake of evaluations, we have attempted grasping all objects in our dataset. For each object, one of the five meshes is selected as the model mesh and the remaining four are used with robot experiments. The model mesh is segmented using k-means clustering and is presented to the user on initial screen. For experiments, we have used $N_C=7$ clusters for each object. The user then selects a region of interest (one of the clusters), after which the model is transferred for grasp generation. Now, with the selected region, a set of $\mathcal{G}^0_{S_\mathcal{X}}$ grasps are generated by our LoCoMo grasp planner and stored. It is worth noting that LoCoMo planner provides ranked list of grasps and when the grasps are transferred, we filter out the kinematically unfeasible ones. For each object category, the experiments are repeated three times for each of the remaining four deformations. Grasps are executed with a force of $50~\mathrm{N}$. Once an object is grasped, we have evaluated the stability of the grasp by conducting the following three tests as in \cite{Bekiroglu2020Benchmark}: (i) lifting test, in which we lift the object $20~\mathrm{cm}$ above the table at the speed of $10~\mathrm{cm/s}$; (ii) rotation test, where we move the manipulator to be in $90-90$ configuration  and rotate the object from $90^{\circ}$ to $-90^{\circ}$ with a speed of $45~\mathrm{deg/s}$; and finally, (iii) shake test, where we shake the object in a sinusoidal pattern with an amplitude of $0.25~\mathrm{m}$ and a peak acceleration of $10~ \mathrm{m/s^2}$. These three tests are performed sequentially in the order mentioned, and if the object slips from the gripper at any point, they are deemed failure and the next test is not performed. The accuracy test is performed by checking if the fingers fall within the chosen region of interest. Sample results are shown in Fig. \ref{fig:FMPointMap}.

Additionally, to compare our method’s performance, we have conducted baseline experiments with CPD \cite{myronenko2010point} and ICP \cite{myronenko2010point}. CPD is implemented using PyPCD library and the ICP using Open3D Python library. The comparison results are summarised in Table \ref{tab:GraspEvaluation}.
\subsection{Discussion}
From analysing the results in Table \ref{tab:GraspEvaluation}, we can clearly see the effectiveness of our method to grasp deformable objects. Out of all object configurations, the proposed FM-based pipeline showed $\mathbf{93.75\%}$ success rate for accuracy tests, as well as an overall success rate of $\mathbf{100\%}$ in lifting objects and $\mathbf{81.25\%}$ of success after rotation and shaking tests are performed. For this set of experiments CPD had a performance of  $56\%$ in grasping the correct region, $68.75\%$ in lifting and $43.74\%$ after shaking and rotation tests, whereas ICP grasped the correct region in $31.24\%$ attempts and had $31.25\%$ success after lifting and $12.5\%$ after the tests. 

Here the poor performance of ICP was to be expected, as it is designed to find rigid alignments and does not deal with deformations. We note that it had some success in grasping objects with little deformation, such as some configurations of the pliers and pipes. In this case, the alignment was not perfect; however, due to our local re-planning it managed some successful grasps. CPD, on the other hand, is equipped to deal with non-rigid transformations, and has been successfully applied to grasp transfer in previous works \cite{Stuckler2014nonrigid, lin2018nonrigid, Rodriguez2018a, rodriguez2018transferring}. However, none of these deal with larger deformations, as CPD is dependant on a good initialisation. We can see that although success in grasping the pliers was almost $100\%$ it failed for most of the other objects, even when it converged to the correct object part. Even with integrated re-planning, when the initial transformation is not good, the grasp fails. The reason is if  the  main transformation  is poor, the final transformation will  be  far  from  the  region  where  the  grasp is originally  planned.

Overall, our FM-based grasping proves to be a strong alternative in the cases where other methods failed due to the larger deformations. Main reason for the high success is because the functional maps are invariant to deformations. Indeed, having high quality functional maps between surfaces is paramount to solving the grasp transfer problem. In fact, from Table \ref{tab:GraspEvaluation}, we see that our method is highly successful for the challenging objects, e.g., Teddy and power Cable. Particularly, only our method managed to transfer grasps with the Teddy Bear object. Besides, the failure case reported with the Pliers object was not due to a poor mapping, but rather because of a change in the object configuration (Fig. \ref{fig:objects}, last row, fourth configuration). As the object was closed, the robot was unable to close the grasp on the left handle, therefore it could not grasp the correct object part. We believe this can be added as an extra check while filtering grasps.

%% file: sections/conclusion.tex
In this paper, we have presented a method for transferring grasps between shapes that have suffered arbitrary deformation. By employing the FM framework, we have shown how the grasps generated on a region of an object can be successfully transferred on to a novel, deformed, instance of the object. FM are particularly robust to large deformations, which was proved by high success rate for a variety of objects with different levels of deformations. When compared with state-of-the-art methods, our method clearly outperformed them in terms of grasp success and grasp region accuracy. 

Apart from optimising for computational efficiency, our approach can be extended in multiple ways. Firstly, we can try to integrate with learning-based approaches to gain semantic understanding of the object and make our method fully autonomous. We can make use of additional sensing information (e.g., tactile) to improve grasp suitability for handover tasks. Finally, it can be extended to perform task-aware grasping by learning object affordances.